# READING ISN'T BELIEVING: ADVERSARIAL ATTACKS ON MULTI-MODAL NEURONS

David A. Noever and Samantha E. Miller Noever

PeopleTec, Inc., Huntsville, Alabama, USA
david.noever@peopletec.com

## ABSTRACT

With Open AI's publishing of their CLIP model (Contrastive Language–Image Pre-training), multi-modal neural networks now provide accessible models that combine reading with visual recognition. Their network offers novel ways to probe its dual abilities to read text while classifying visual objects. This paper demonstrates several new categories of adversarial attacks, spanning basic typographical, conceptual, and iconographic inputs generated to fool the model into making false or absurd classifications. We demonstrate that contradictory text and image signals can confuse the model into choosing false (visual) options. Like previous authors, we show by example that the CLIP model tends to read first, look later, a phenomenon we describe as reading isn't believing.

## 1. INTRODUCTION

The blending of text and imagery in learning models introduces a touchstone for exploring multi-modal tasks [1-7]. This research explores the novel possibilities for contradictory inputs and using synaesthesia as a metaphorical probe or method to understand multi-modal neural attacks. The Greek roots for "synaesthesia" translate to "perceive together".

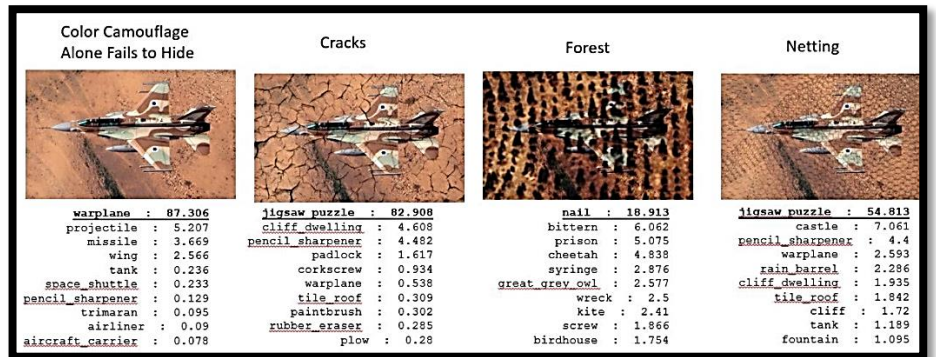

*Figure 1. Texture Multi-Modal Attacks on Image Classifiers. An analogous sensory cross-over between convolutional neural networks which prove more sensitive to texture (touch) than shape (vision) compared to humans.*

The experience of overlapping cognitive responses creates a potential for disagreement or classification dissonance [6]. In the present study, "recognizing while reading" serves as a metaphor for recently discovered contradictory inputs as attackers might present them to multi-modal neurons. To blend reading and object recognition, Open AI introduced their CLIP model (Contrastive Language–Image Pre-training) [1]. CLIP includes a traditional convolutional neural network (CNN) for image classification but adds a language transformer (encoder-decoder architecture) to associate text labels to the object. The authors trained CLIP to recognize and classify the literal, figurative, and symbolic versions of a given object [1]. By combining text and imagery, CLIP offers a way to probe the limits for fusing a realistic image (e.g. apple), a figurative version (e.g. line drawing), and its language representation (abstraction, e.g., "Granny Smith") in text. What happens if the three meanings do not match?

We explore interrupting the model's expectations for consistent sensory inputs. In Figure 1, we show an image-only example of a texture-based probe for convolutional neural networks, one which highlights the adversarial risks of fielding an image classifier with a backdoor method to alter its decision-making. In this case, the carefully designed iteration of different textures can alter the image class without disguising its overall shape in a way that would fool the human visual system. Machine learning at fine-scale is not only shape-dependent but texture-sensitive compared to human vision. Creating stealth methods for hiding objects in plain sight is akin to building stealth aircraft with sharp, non-reflecting edges to fool radar detection. Figure 2 shows that the classic image-based sensitivity for texture over shapes, but in the CLIP model, the text label can over-ride the texture and the image shape. This style of adversarial attack gets additional options when the CLIP model reads the label, sees the texture, and classifies the shape.

The CLIP authors [1] note that their neural bridge opens up more complex forms of visual recognition that traditionally elude standard object classifiers [1-3].

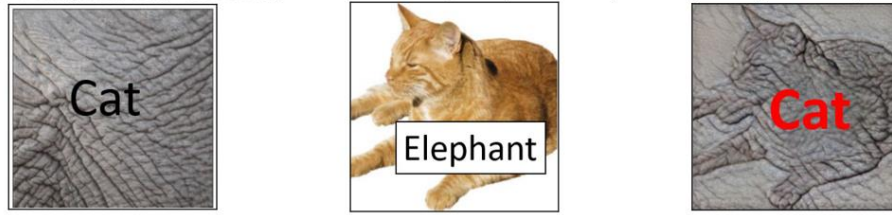

Figure 2. Illustration of CNN sensitivity to texture in ResNext, but CLIP sensitivity for text over texture.

They note that previous work has struggled to understand visual tasks involving geographic regions, facial and emotional expressions, religious iconography, and animal behavior (among others), but their multimodal neurons in CLIP offer a deeper conceptual framework [1]. Their CLIP investigations highlight abstract instances of counting, art appreciation, and counterfeit image detection. Goh, et al. [6] examined CLIP to uncover what they refer to as fallacies of abstraction, mainly using images of text as probes of the model's abstractive layers as well as ways to exploit its reductive classifications in a game of absurdities and foolish decisions. One example is draping a poodle (realistic image) in dollar signs (figurative money) to generate a CLIP prediction of poodle as a piggy bank (abstractive "finance"). Using images of text to fool an image classifier with multimodal neurons present what the authors [6] called "typographic attacks".

In response to these misclassifications, the original CLIP authors indicate the significance of any attacks hinge on whether the model is fielded in the wild [8]. A text attack in a lab renders it more a theoretical notion, one which might self-heal if presented with the right type and size of training data. But if one imagines the model never gets real-world inputs, why go to the trouble to examine its many biases like associating race with animals or ethnicities and gender with violence? Goh, et al. [6] compare their typographical attacks to well-known adversarial patch attacks [9-11], the significance of which can mask traditional image classifiers for people on CCTV cameras or school buses in traffic. They further note that outside the laboratory setting, many real-world objects routinely combine objects with contradictory text (e.g. suggestive product logos) and thus as a general notion, present challenges to combining recognition models with reading skills. How humans learn that a Volkswagen brand car is not an actual "bug" seems not to depend on reading ability as much as reasoning with object recognition. Similarly, the text in an airline logo represents a genuinely new form of camouflage to a model that reads before it decides if it's an airplane or not. One can also envision positive applications of this reading ability in areas like image watermarking, copyright, or trademark protection [12]; these cases benefit from a contradictory textual input that might over-ride the object identification. If one expands the camouflage example, it is possible to imagine multiple ways to disguise an airplane [13]. Previous workers have noted the ability to fool convolutional neural networks with texture alone that over-rides shape in standard object models [14]. In this way, putting an animal logo, national flag, or foreign typography on an airline might render its fleet invisible to multi-modal neural models. The purpose of the present research is to explore and expand on the basic tension between contradictory inputs to an algorithm weighted across the spectrum of literal, figurative, and abstractive reasoning skills.

## 2. METHODS

The approach to shifting the CLIP decision boundary is necessarily experimental (Figure 3). Since the model is publicly available, this attack style benefits from knowing the model lineage, its CNN, and Transformer architecture. The attacks are thus "white-box" since the "black-box" probes to identify the model or weights can be largely skipped [15]. However, in the CLIP case, the larger questions explored here center around the model's dominant reading skill which over-rides the visual input. Unlike the pen-and-paper attack shown by Goh, et al. [6], the complexity of these attack styles ranges from simple to multi-step probes to fool the model.

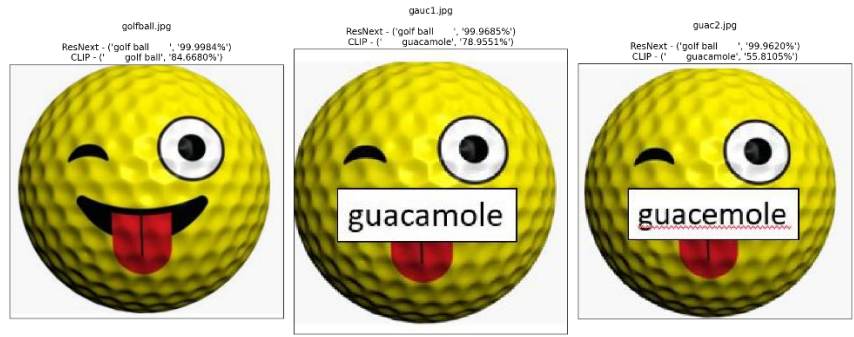

*Figure 3. Basic Typographical attack on ambiguous objects and misspelled labels. CLIP correctly characterizes the stylized golf ball but concludes incorrectly with text overlays even with misspellings.*

| Categories | Attack Variants | | | | | | |
|---|---|---|---|---|---|---|---|
| **Typography** | O-O-V | Synonyms | Foreign | Spelling | Text Flood | Size | Orientation |
| **Conceptual** | Image Font | ASCII Art | CAPTCHA | | | | |
| **Imagery** | Masked | Word Image | Logos | Paint-by-numbers | | | |
| **Figurative** | Dot Art | Skeletonized | | | | | |

*Table 1. List of possible contradictory or confusing input types to test multi-modal classification models. OOV refers to out-of-vocabulary text inputs.*

Table 1 summarizes some basic typographical, conceptual, and iconographic attack categories investigated. Through careful construction of examples, we generate a basic template for the image classification alone, followed by various contradictory inputs that might alter the decision boundaries. For example, forcing the model to "read first, look later" offers nuanced attacks that feature foreign languages, synonyms, spelling differences, background text, font size, and orientation of both the object (horizontal bar) and text (vertical or rotated). In addition to these typographical attacks, we constructed new categories for conceptual and figurative attacks. By embedding images into the font, we further explore the feature weights behind what represents text or imagery. This style of attack might be comparable to traditional cases in the wild, such as ASCII art and CAPTCHAs ("I am not a robot" images and text mixed [16]). Another use case encountered often in the wild for stylized text includes brand logos and highly abstracted visual representations. Finally, we construct examples to detect if the model can complete an otherwise incomplete figurative representation, similar to connect-the-dots or paint-by-number examples.

## 3. RESULTS

To compare these results with previous work, we adopt the template for the image first, followed by the assorted variations in the text to fool the CLIP model [17]. Each sub-section describes the motivation behind the experiment and demonstrates the attack's success using the probable class assignments for both the image-only model (ResNext [18]) and the multi-modal CLIP [1] classes. The classes for these experiments derive from the 1000 ImageNet dictionary [19], one which includes many natural world examples for animals (including sub-species), plants, and insects. The appendix show other iterations on a theme for examples with real-world significance in facial recognition for celebrities and classically mislabelling a "stop sign" as a "stop watch".

*3.1. Typographical Attack: Near Match.* This attack reproduces the previous methods [6] to overlay text ("bee") on a non-matching image ("Granny Smith apple"), a contradiction that forces the model to choose the higher abstracted version which is the text label ("bee"). The variant shown in Figure 4 illustrates further that even though the model reads the label, the attacker can choose an out-of-vocabulary word ("beetle") which is a near match, and the model again will choose the higher abstracted member ("bee") in its dictionary. To show the versatility of this attack style, the preference from text label over the imagery is illustrated by the superseded "bee vs. apple" classification symmetrically, in one case with greater than 98% confidence that an apple with the text label "bee" is indeed a bee. The CLIP response to minor alterations in a text (like misspelled labels already in its vocabulary) however does not yield the same result. Rather a single letter (alteration of "Granny" to "Grannie"), produces an image-dominated classification.

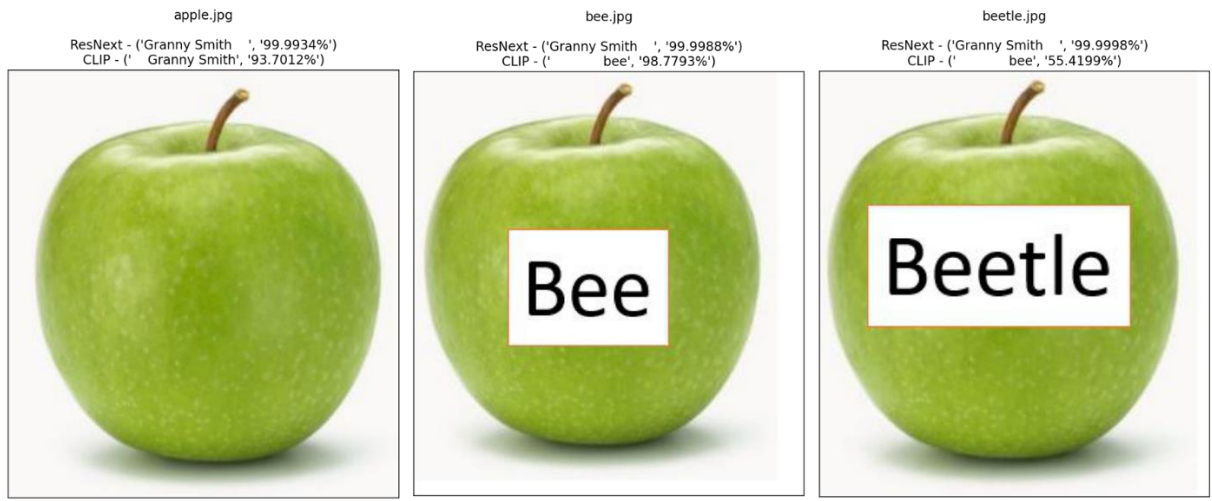

*Figure 4. Typographical attack with out-of-vocabulary label that the model brings back into its own vocabulary*

*3.2. Typographical Attack: Text Size.* In Figure 5, this attack illustrates the influence of text size on misclassifications. For the Model-T image, a smaller overlay of "bike" in text reduces the certainty for CLIP but does not alter the top choice. When we enlarge more specific in-vocabulary text labels "mountain bike" and "oxcart", the CLIP model accepts the text over the underlying image. We also vary some of the labels with all capital letters and decorated font styles without significant effect in this attack.

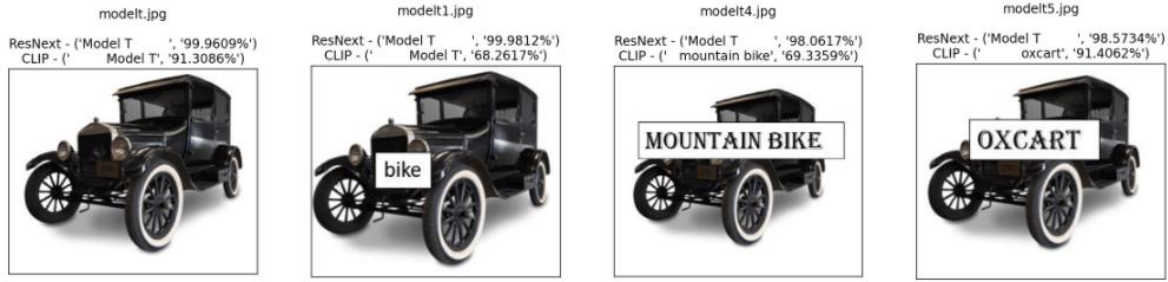

*Figure 5 Typographical Attack using Font Size*

**3.3. Typographical Attack Per Letter: Text Font.** In Figure 6, this attack introduces the text as all capitals and embedded figurative imagery embedded in the letters. The target category is flower parts such as "hip" within ImageNet, which in this case, only ResNext scores correctly. CLIP neither understands the words or the image. This variant illustrates the influence of word cases and introduces a secondary capability to confuse CLIP with a contradictory word and image per letter.

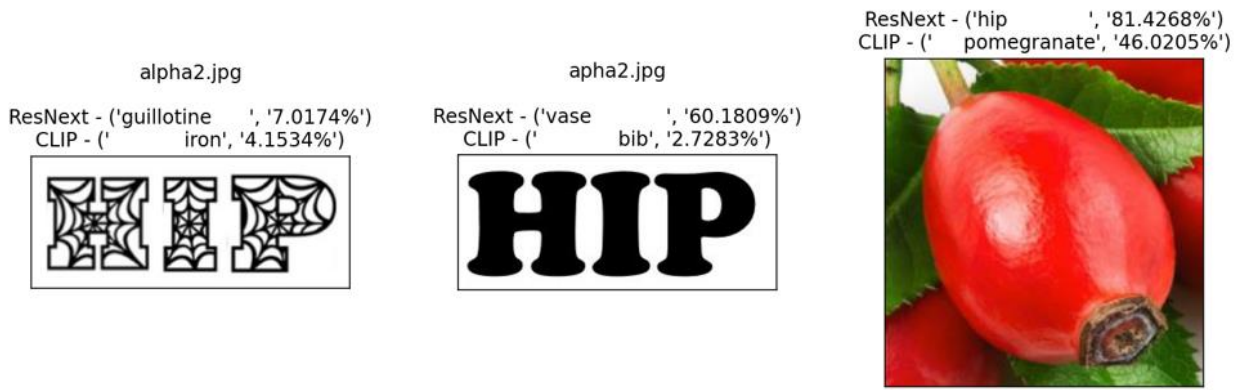

*Figure 6. Typographical Attack Per Letter and Font*

**3.4. Patch Attack: Image Font.** In Figure 7, this attack directly embeds an ImageNet class ("spider web") into the font face. This style of attack [10] shares some features with the traditional Patch Attack, which masks the underlying word with an image. These variants reverse what Goh, et al. [6] pointed out as the CLIP model overweighting text reading compared to image recognition. Rather, this image font attack illustrates the model's ability to see the embedded image ("spider web") and not see the word ("alphabet").

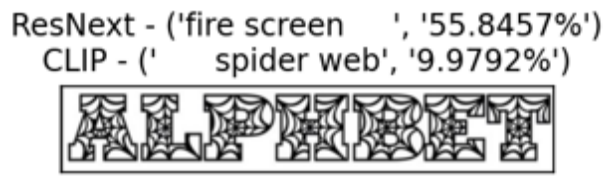

*Figure 7. Patch Attack using Image Font*

**3.5. Flood Attack: Image Identification Inside Lettering.** This attack illustrates a multi-stage attack, which both includes letters in the image class by definition ("crossword puzzle"), but further reinforces that with a flood of alternative readable labels. In the absence of conflicting inputs, the attacker can force CLIP to misclassify the image (an empty "crossword puzzle") because of the overriding text ("meat loaf") label. However, in the second instance, with a background of alternative labels, the size of the typography introduces another CLIP vulnerability. This effect in Figure 8 might be compared to a classic signal-to-noise problem. In the presence of background text, the model requires a dominant font size before flipping its decision from "crossword puzzle" to the attacker's choice, "meat loaf".

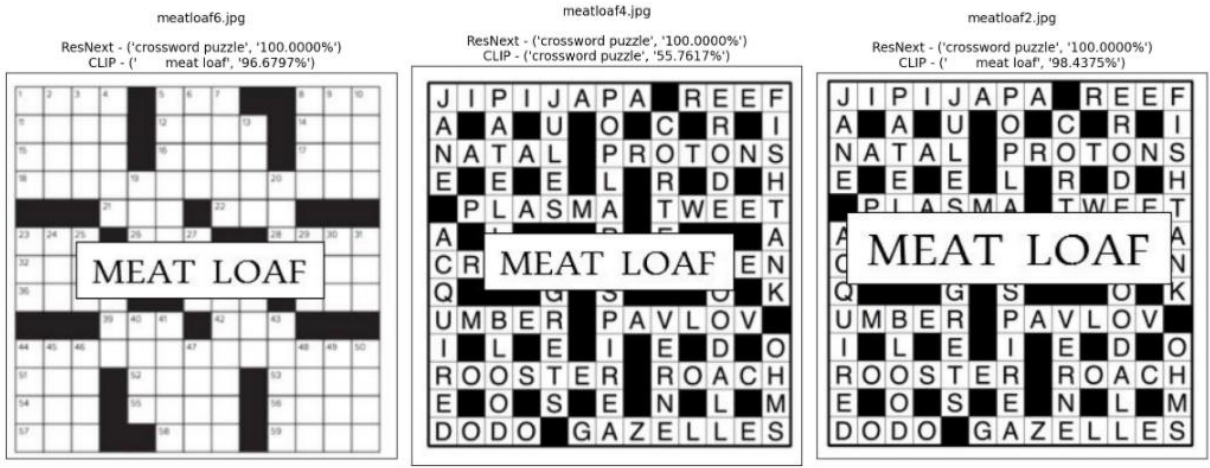

*Figure 8 Flood Attack: Image Identification Inside Lettering*

***3.5. Flood Attack: Image Identification Inside Lettering.*** This attack illustrates CLIP sensitivities to where the text label is shown within the image and further explores embedding images within and around letters as decorative attacks. In contrast to a font built on an image (the "spider web" alphabet in 3.4, the patch attack, this case shows a more peripheral placement. In Figure 9, the baseline image is "garden spider" from ImageNet, the written label is the classic story twist in the book "Charlotte's Web", where the spider spells out words. In this case, CLIP misses those words and with 81% confidence recognizes the image over the typography. However, when the word "spider" is decorated with web elements exterior to the letters, the CLIP model ignores the word and focuses on the tiny dot ("tick") above the letter "i".  A further variant on this decorative or flood attack introduces another literal conclusion from CLIP. In the last case, the out-of-vocabulary shortened term "web" does not trigger CLIP to see a spider any longer (despite the spider web iconography). Instead, the partial term "web" triggers the literal "web site" classification.  It is worth noting that without the ability to read words, ResNext similarly focuses on elements of the word to derive various outlined classes like "sunglass" and "loupe" but with low confidence.

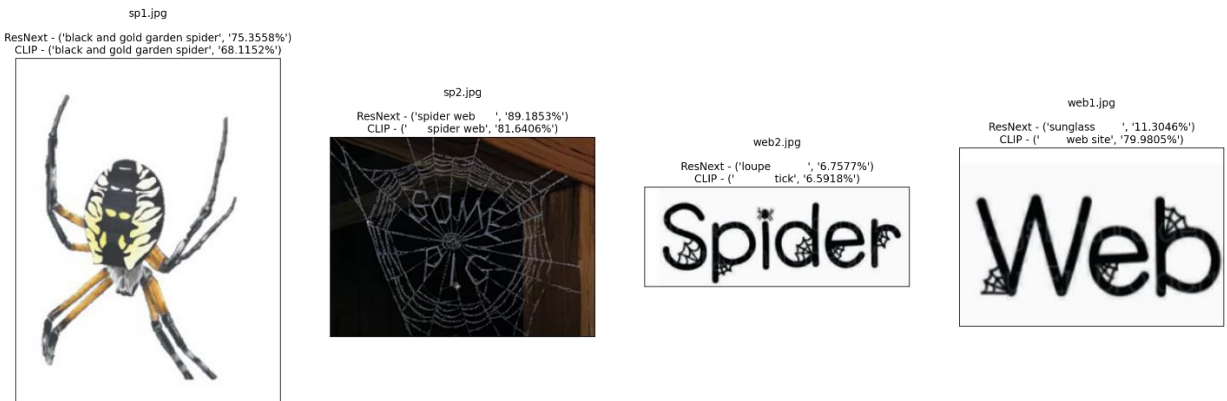

*Figure 9 Flood Attack: Image Identification Inside Lettering*

***3.6. Rotation Attack: Image Identification with Symmetry Defining Label.*** In Figure 10, this attack introduces what might be commonly understood as "world knowledge" testing. When CLIP sees the object "horizontal bar", it recognizes it as such whether it is vertical or horizontal. One might conclude this rotational invariance follows from the training style of showing images in different orientations to increase the model's ability to generalize, but in this case, leads to a contradiction or lack of world knowledge. The consequences of not understanding that things fall ("gravity") or flow downhill ("water") leads to interesting methods to uncover the depth of a model's true comprehension (vs. memorization or mimicry). One test of this rotational invariance is to incline the bar at forty-five degrees, which indeed does push CLIP to describe not a bar but a ladle. This ambiguity could also serve as an endorsement of multi-modal neurons since one could presumably impose the correct label in text and reassert the correct class via the secondary input. In our case, we describe the typographical attack with an in-vocabulary text label "joystick" which given the low confidence from the image itself, the model succumbs to the literal text.

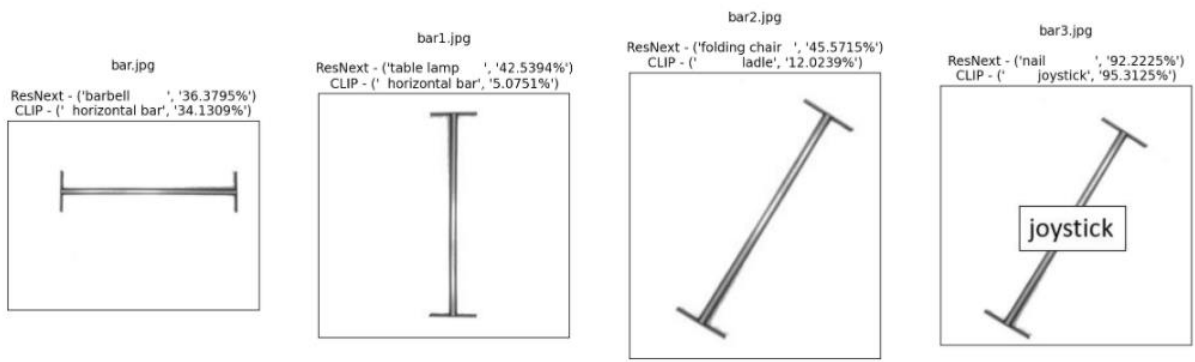

*Figure 10 Rotation Attack: Image Identification with Symmetry Defining Label*

***3.7 Steganography Attack: Hiding Letters Inside Image.*** In Figure 11, this attack shares some design features with the previous crossword puzzle example (3.5). Their common theme is the integral nature of text and image together as a test for multi-modal decisions. As part of ImageNet classes, the "space bar" on a keyboard or typewriter necessitates the coincident appearance of background letters like a crossword puzzle. In the base case, ResNext recognizes the space bar where CLIP recognizes the whole typewriter keyboard. By superimposing the text "space bar" in all capitals, the reading portion of CLIP adds to the confidence for the correct class. Interestingly, when the font spelling out "space bar" resembles the usual keys in both their background color, font type, and relatively wider spacing, the CLIP model revers to a lower confidence decision of a keyboard. To test if the letters themselves can trigger a multi-modal neuron, we further include the letters spelling "space bar" onto the keys themselves ("A-K"). In this case, the model potentially has the correct typographical and the correct unchanged physical bar in the same view. The result is that the key placement of the label gets lost to CLIP and the class reverts to the lesser confident "keyboard". This outcome mirrors what attack 3.5 showed, where the font size "meat loaf" had to be sufficiently uncluttered or large to change a decision boundary. Altering the spelling with a partial "pace bar" by removing the leading "s" neither changed to class nor the confidence for a "keyboard" assignment.

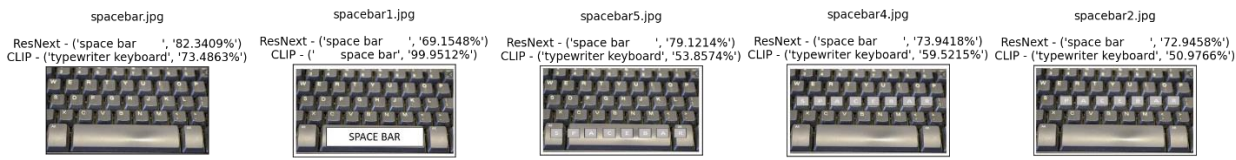

*Figure 11 Steganography Attack: Hiding Letters Inside Image*

***3.8 Homonym Attack: Conceptual Puns.*** In Figure 12, this attack introduces a physical pun between a human and corn "ear" and then sways the model to decide by imposing a typographical attack. The size and background of the typography further define which direction the model will predict. This pun between two competing concepts might at first appear to be a minor consequence of the broader "read first, look later" dominance. But one might argue that introducing logical contradictions between senses, words, or concepts represents a core human capability for humor or sarcasm. If puns are not a particularly sophisticated form of a joke, they do offer some insight into how more complex concepts might be introduced to machine learning models. For instance, an unexpected turn of events in the wild may represent just another story-telling technique that appears often as contradictory inputs, either to one's expectation of events or the actual sensory expectations. The Appendix shows further iterations on concepts, puns, and labels that continue the theme using the singer, Mr. Loaf, meatloaf, loafers, and French loaf. Notable findings there include the label dependence on font color (red letters fail to shift where black letters succeed). The spacing between the terms "meat" and "loaf" does not alter the labels when superimposed over "French loaf". Remarkably, CLIP recognizes celebrities like singer Mr. Loaf as actual meatloaf.

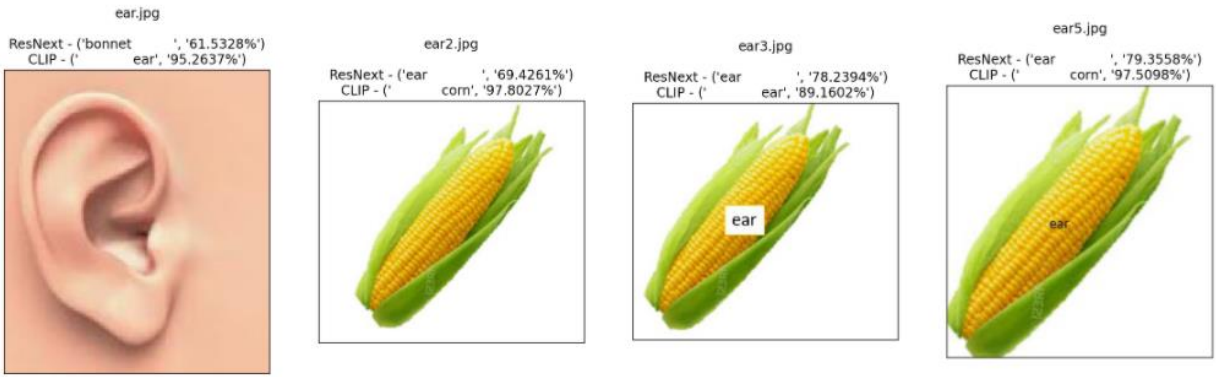

*Figure 1 Homonym Attack: Conceptual Puns*

***3.9. Clustered Typographical Attacks: Related Text for Unrelated Concepts.*** In Figure 13, this attack builds on the ImageNet hierarchy of labelling conventions that center on the fragmentary "bag". The baseline image is correctly identified by CLIP as "purse" and ResNext focuses on the "buckle". This class however is not interchangeable as a concept with "bags" used for different functional purposes, like "sleeping" or "punching", or unlikely purse materials like "plastic". When the typographical attacks for each combination over-ride the actual image, CLIP follows the usual pattern of deferring to the literal text. If labelled with the label fragment "bag" alone, however, the multi-modal neurons give the correct label as "purse" which also agrees with putting the label itself "purse" over the image.

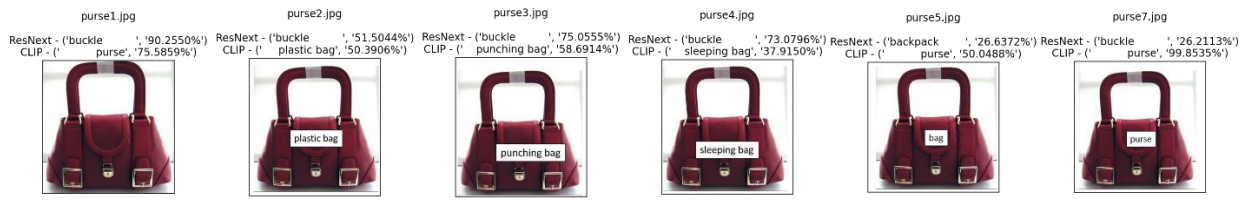

*Figure 13 Clustered Typographical Attacks: Related Text for Unrelated Concepts*

***3.10. Figurative Logo with Text.*** In Figures 14-15, this attack introduces a simple brand logo for an airline. As discussed in the introduction, the fusing of text, icons, and images in brand logos likely present real-world instances where a confused classifier might shift its decision boundaries on trivial changes in text color, font size, or other parts of the attacker's toolkit. The ability to disguise an airplane as a kangaroo (Qantas) because of its logo serves as one initial example of adversarial attacks beyond just the academic typography. We illustrate the potential multi-modal confusion in a time series of changed logos for Hawaiian Airlines. With the text plus image as a baseline, CLIP reads the text absent any clues for an airplane in imagery and correctly classifies the family of concepts to "plane". Without the text triggers, CLIP neither recognizes the underlying "flower" iconography, either in the background or in the hair. By only changing colors in the iconography, this attack shifts CLIP and ResNext into entirely different domains, such as "sea anemone" vs. "web site". In many ways, brand logos force both a human and a multi-modal model to dismiss their literal tendencies in favor of a more imaginative interpretation. The logo for "golden arches" represent the letter "M", but under both text and image attacks can shift the CLIP

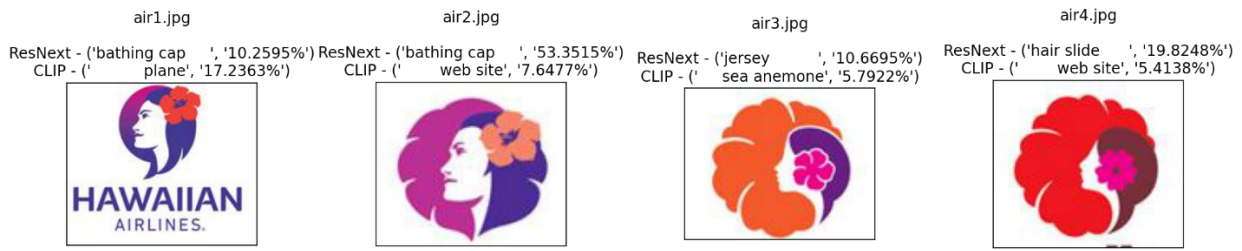

*Figure 14 Figurative Logo with Text*

decision to classify "pizza".

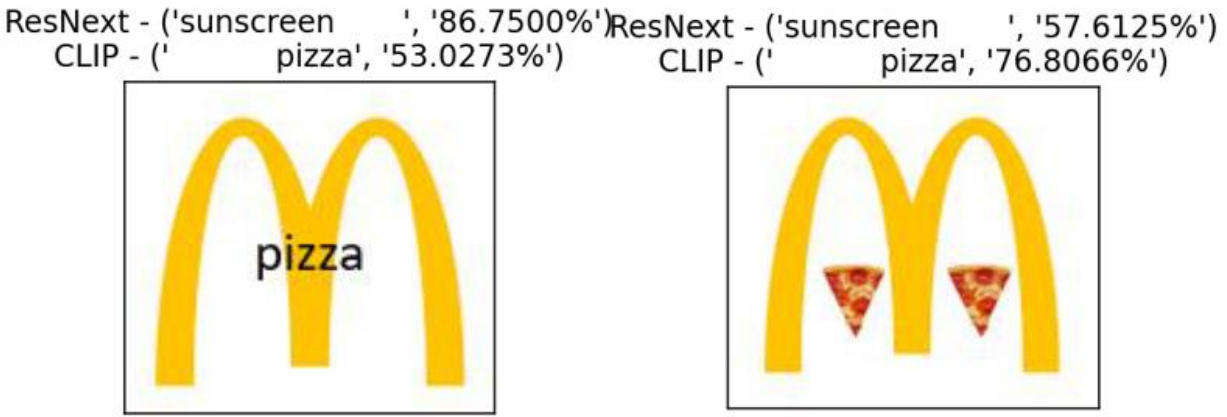

*Figure 15 Figurative Logo with Text*

***3.11. Font Complexity.*** This attack differs from previously embedding contradictory imagery into the text font (3.3, 3.4). In Figure 16, the text font itself and word meaning coincide. This illustration also mimics the basic CLIP capability to recognize the realistic image (garden "maze"), the figurative variation (puzzle "maze"), and the abstract representation in (font-dependent) language. As the font complexity increases and its letters more closely resemble the object it describes, the farther the CLIP model drifts from understanding the word and jettisons the image class. More simple lettering enables the model to read and associate the "maze" with the concept. As the font starts to resemble the object itself, the model incorrectly classifies the word representation as "web site". This style of typographical attack with increasing font complexity but increasing realism towards the actual object that the word symbolizes may present a likely candidate for confusing CLIP in the wild. Instances of brand logos using typeface to identify both the word and object were explored in 3.10.

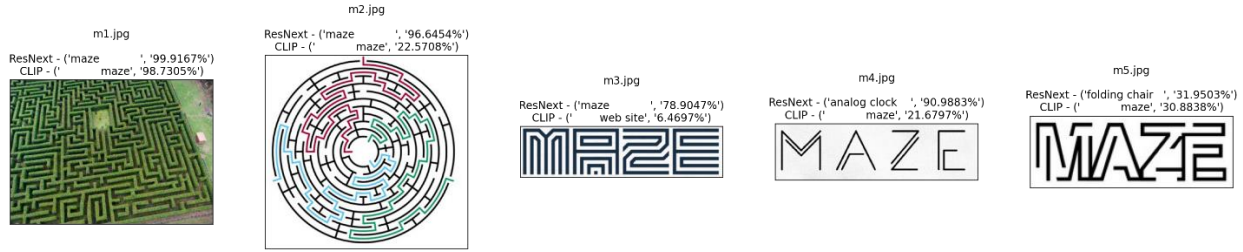

*Figure 2 Font Complexity*

***3.12 Conceptual Complexity or Iconographical Attack.*** This attack resembles the initial CLIP association between related symbology for dollar signs and finance. The authors [1] mention religious iconography as one compelling instance where previous image-only models might fail but a combined or multi-modal approach might deepen recognition and understanding. In this case, we explore the model's understanding of medical symbology. In Figure 17, the baseline image (featuring two intertwined snakes, caduceus) is used incorrectly (the symbol of Hermes). CLIP associates the symbol to medical fields by classifying it as a "stethoscope" where ResNext recognizes the shape itself as a "hook". The first variant is the Rod of Asclepius, which is the better historical symbology [20] but loses its CLIP association with anything medical related ("hook"). This example suggests the training data is incorrect but learned correctly by CLIP. More medical symbology further defines the family of clustered labels, with a circle plus sign associated with ambulance in CLIP but not ResNext. Similarly, CLIP does not associate iconographic images of DNA with anything medical ("abacus"). To underscore the possible typographical features, the final example imposes an antonym ("poison") to overlay the traditional medical snakes, but CLIP changes its classification from stethoscope to "medicine chest". In this case, the associated field might be characterized as correct, but the typographical overlay suggests danger or perhaps the need for medical care rather than the care itself.

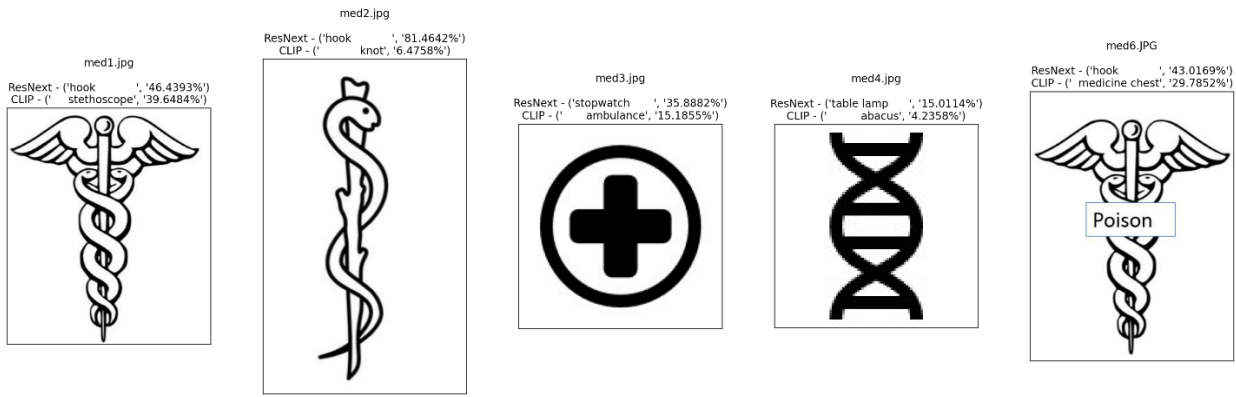

*Figure 3 Conceptual Complexity or Iconographical Attack*

***3.13. Iconographic Attack.*** This attack was first demonstrated by the CLIP authors to show that the overlap of a dollar sign with a puppy generates the class for finance and a piggy bank. In Figure 18, this variant however again changes the underlying sentiment of the image compared to the typography or iconography. As noted by Goh, et al. [6], the dollar sign icon communicates an association with finance as a baseline case. When imposed on images more aligned with poverty and homelessness, the model shows some of the traditional cultural biases such as race, ethnicity, gender, and socioeconomic status. Putting dollar signs over the text of "hungry, need help" shifts CLIP to again label the image as "piggy bank". We further explore the mixture of images and text in the same finance-related group with a dollar bill (CLIP="wallet"), "hunger" sign (CLIP="doormat"), and a beggar's cup with a dollar sign and bill in the same image (CLIP="piggy bank"). This example sequence highlights how contradictory inputs favor the symbology ($) itself over text and images. We further uncover what some associations might lead to mixed messaging, similar to shifting the sentiment score in a surprising way given the underlying image class.

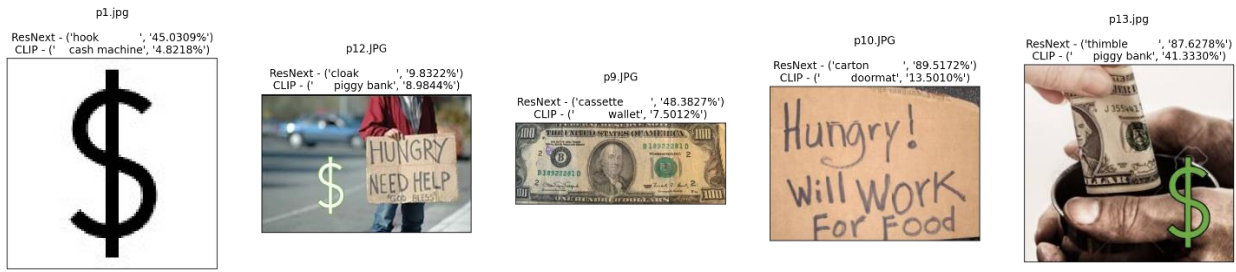

*Figure 18 Iconographic Attack*

## 4. DISCUSSION AND CONCLUSIONS

The reported experiments highlight interesting conflicts between text and images as inputs to a multi-modal neural model called CLIP. The text transformer dominates the image CNN for most classifications examined here, a finding that confirms what Goh, et al. first noted [6]. The experiments explore the fuzzy boundaries between this "reading first, look later" approach. The results show that the model reads between the lines for attacks based on near text matches that are misspelled, synonyms, or otherwise typographical iterations. The results further enhance the limits of images combined with the text itself, such that the model misclassifies for different font sizes, shapes, or patterns. We add further examples to illustrate the non-typographical attacks, those that alter the iconography or concept itself. This approach shares some similarity to a human sense of humor itself when it introduces puns and other unexpected combinations of meaning. We anticipate that these attacks only scratch the surface of the attacks possible with mixed messaging between image, text, and icons.

### ACKNOWLEDGMENTS

The author would like to thank the PeopleTec Technical Fellows program for encouragement and project assistance.

## Appendix: Example Experimental Results

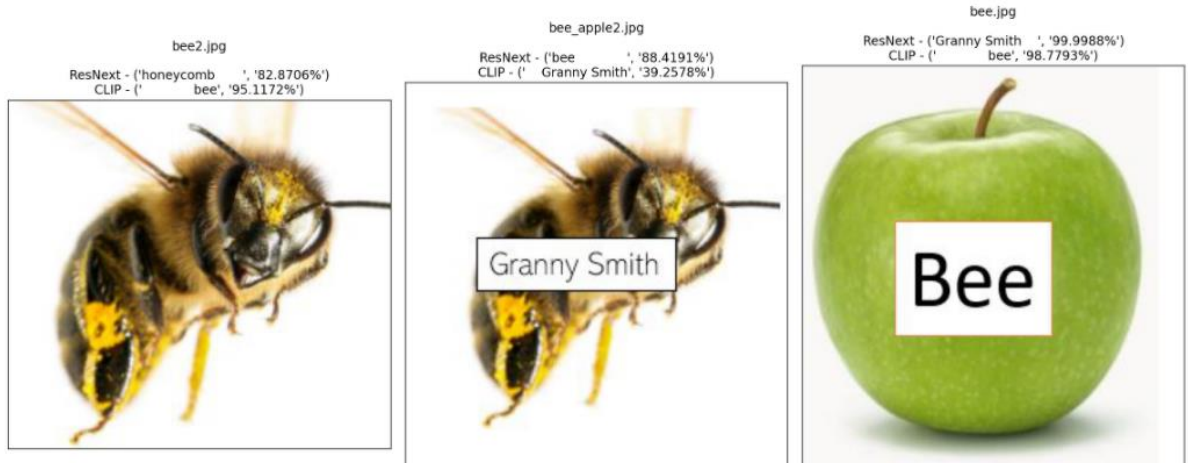

*Figure 4. Symmetric labelling overrides imagery with high confidence*

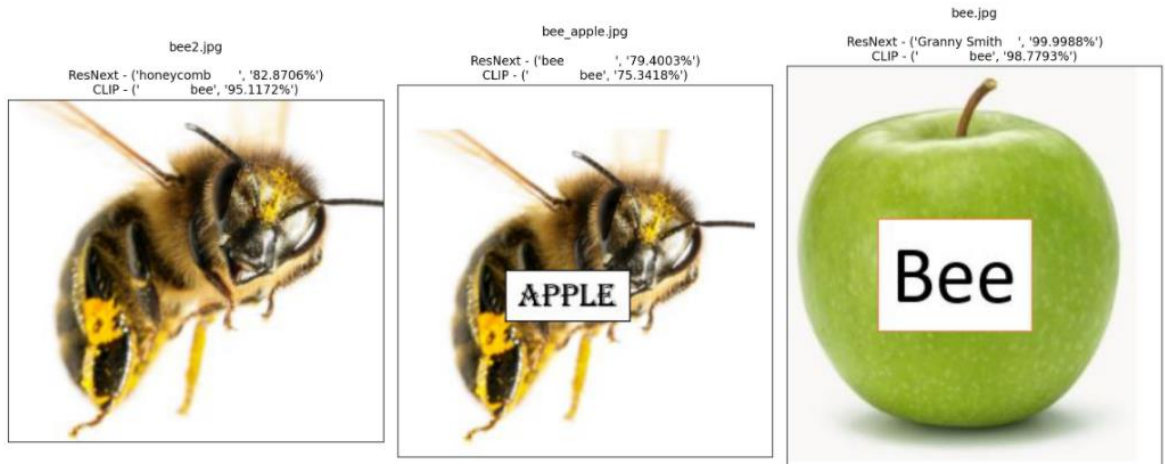

*Figure 20. Symmetric labelling with an out-of-vocabulary label*

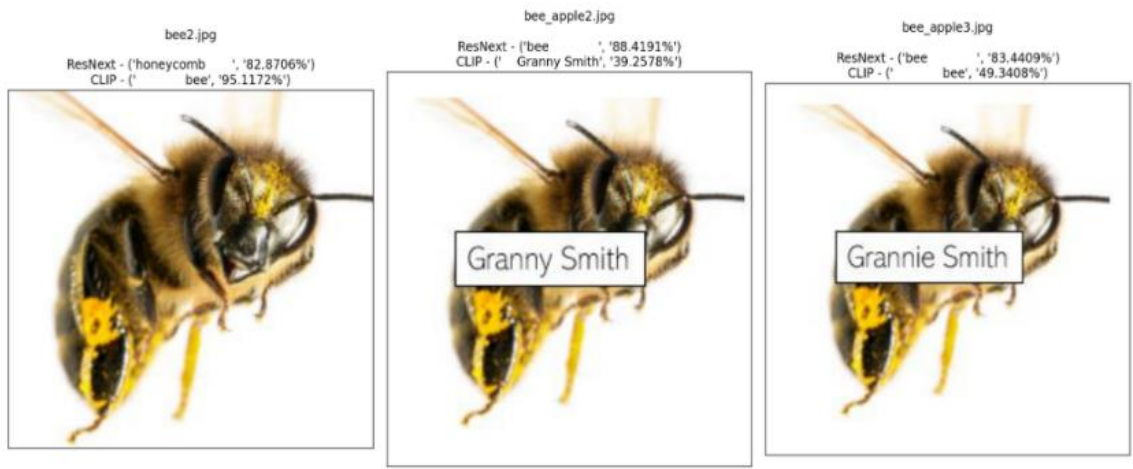

*Figure 5 Typographical attack with single letter misspelling*

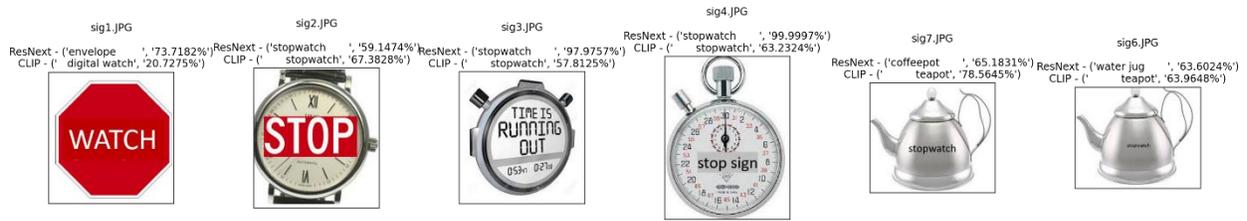

*Figure 22. Iterations on a theme with watch, stopwatch and stop sign*

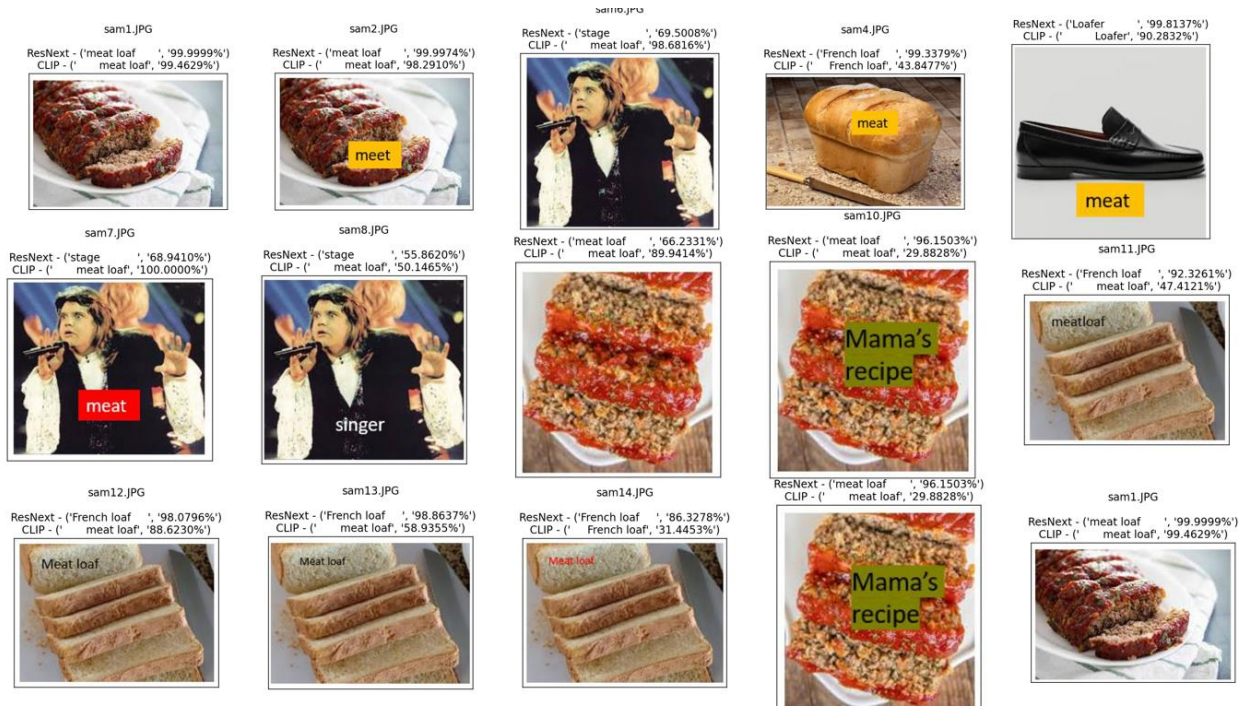

*Figure 23 Iterations on a theme with the singer Mr. Loaf, a loafer shoe, actual meat loaf and French bread with confusing labelling.*

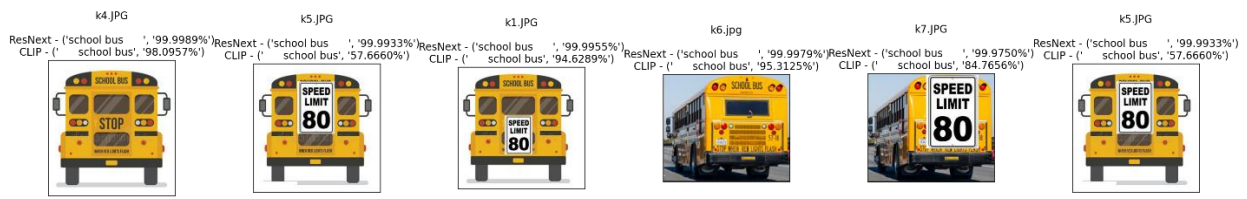

*Figure 6. CLIP Recognizes Object over Text for Street Signs*

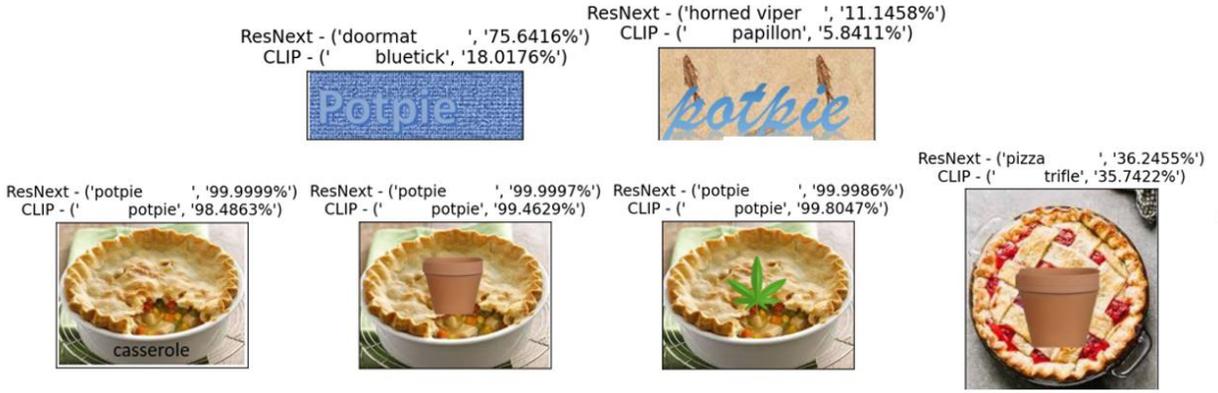

Figure 7 Typeface Texture Attacks